\pdfoutput=1

\documentclass[11pt]{article}

\usepackage{EMNLP2023} 

\usepackage{times}
\usepackage{latexsym}

\usepackage[T1]{fontenc}

\usepackage[utf8]{inputenc}

\usepackage{microtype}

\usepackage{booktabs}
\usepackage{graphicx}
\usepackage{CJKutf8}

\title{XuanYuan 2.0: A Large Chinese Financial Chat Model \\ with Hundreds of Billions Parameters}

\author{Xuanyu Zhang\textmd{,} Qing Yang \textmd{and} Dongliang Xu \\
\\
Du Xiaoman Financial}

\begin{document}
\maketitle
\begin{abstract}

In recent years, pre-trained language models have undergone rapid development with the emergence of large-scale models. However, there is a lack of open-sourced chat models specifically designed for the Chinese language, especially in the field of Chinese finance, at the scale of hundreds of billions. To address this gap, we introduce \textbf{XuanYuan 2.0} (\begin{CJK}{UTF8}{gbsn}轩辕\end{CJK} 2.0), the largest Chinese chat model to date, built upon the BLOOM-176B architecture. Additionally, we propose a novel training method called hybrid-tuning to mitigate catastrophic forgetting. By combining general-domain with domain-specific knowledge and integrating the stages of pre-training and fine-tuning, XuanYuan 2.0 is capable of providing accurate and contextually appropriate responses in the Chinese financial domain.

\end{abstract}

\section{Introduction}

In recent years, pre-trained language models have witnessed rapid development.
Broadly speaking, they can be categorized into three main architectures: the Encoder architecture represented by BERT \cite{devlin2018bert}, the Decoder architecture represented by GPT \cite{radford2018improving}, and the Encoder-Decoder architecture represented by T5 \cite{raffel2020exploring}. Each architecture has its unique characteristics and advantages, catering to different NLP requirements.

The GPT series, with GPT-4 \cite{openai2023gpt4} being the latest addition, has gained considerable attention due to its remarkable performance in natural language generation tasks, including dialogue generation. The ChatGPT \cite{openai_chatgpt} model, in particular, has impressed researchers and practitioners with its ability to generate coherent and contextually relevant responses in conversational settings. As a result, the GPT series has become a focal point of research and development in the NLP community.

Moreover, the emergence of large-scale pre-trained models has further fueled the advancements in language modeling. 
Models such as OPT \cite{zhang2022opt}, BLOOM \cite{scao2022bloom}, and LLaMA \cite{touvron2023llama}, with parameter sizes reaching billions, have recently been open-sourced, enabling researchers and developers to explore the potential of these massive models. These models have demonstrated superior performance on various tasks, pushing the boundaries of what is possible in NLP.

\begin{table*}[!htb]
\centering
\begin{tabular}{p{5cm} l l p{6cm}}
\hline
\textbf{Model} & \textbf{Type} & \textbf{Parameter}  & \textbf{Corpus Content}\\
\hline
FinBERT~\cite{araci2019finbert}
& PLM
& 110M
& News filtered by financial keywords 
\\
FinBERT~\cite{yang2020finbert}
& PLM
& 110M
& Corporate Reports, Earnings Call Transcripts, Analyst Reports
\\
Mengzi-BERT-base-fin~\cite{zhang2021mengzi}
& PLM
& 110M
& News, Analyse reports, Company announcements
\\
FinT5 \cite{lu2023bbt}
& PLM
& 220M, 1B
& Corporate Reports, Analyst Reports, Social media and Financial News
\\
\hline
XuanYuan 2.0
& ChatLM
& 176B
& Corporate Reports, Analyst Reports, Social media and Financial News
\\
\hline
\end{tabular}
\caption{Comparison of different financial language models.}
\label{tab:typicalplm-corpus}
\label{tab:related}
\end{table*}

While the general-purpose large models mentioned above have garnered significant attention, the importance of domain-specific models cannot be overlooked. In many domains, the distribution of language and the specific linguistic nuances require models that are fine-tuned or specifically trained for that particular domain. Consequently, a range of domain-specific large models has been proposed to cater to the unique needs of various fields. For example, BioBERT \cite{lee2020biobert} and PubMedBERT \cite{gu2021domain} are proposed for the biomedical field, and BloombergGPT \cite{wu2023bloomberggpt} are proposed for financial scenarios. These models have shown promising results in their respective domains, leveraging the domain-specific knowledge learned during pre-training.

Within the Chinese financial domain, there has been considerable progress in the development of pre-trained language models. Researchers have introduced models such as FinBERT \cite{araci2019finbert,yang2020finbert,liu2021finbert}, Mengzi \cite{zhang2021mengzi}, and FinT5 \cite{lu2023bbt}, which have been tailored for financial text analysis and understanding. These models, though valuable for certain applications, have parameter sizes below one billion, limiting their ability to handle the increasing demands of the Chinese financial NLP landscape. 
As the volume of financial data and the complexity of language usage continue to grow, there is a pressing need for more powerful models that can effectively process and understand Chinese financial text.

Despite significant advancements in chat models, there is currently no open-sourced chat model at the scale of hundreds of billions specifically designed for the Chinese language, let alone in the field of Chinese finance.
To address this gap, we propose \textbf{XuanYuan 2.0} (\begin{CJK}{UTF8}{gbsn}轩辕\end{CJK} 2.0), the largest Chinese chat model to date, based on BLOOM-176B. XuanYuan 2.0 not only surpasses its predecessor, \textbf{XuanYuan 1.0} (\begin{CJK}{UTF8}{gbsn}轩辕\end{CJK} 1.0), which achieved first place at the leaderboard of CLUE classification in 2021, but also addresses the need for a large-scale chat model specifically designed for the Chinese financial domain.

Furthermore, domain-specific language models and chat models impose higher requirements on data distribution and training approaches compared to general-domain models. Domain-specific models need to capture the unique linguistic characteristics, terminologies, and contexts of a particular field to achieve optimal performance. However, training these models solely on domain-specific data may lead to catastrophic forgetting, where the model loses previously learned knowledge from the general domain, impacting its overall performance. To mitigate this issue, we propose a novel training method, hybrid-tuning, that combines the stages of pre-training and fine-tuning. 
By integrating the two stages, our approach guarantees that fine-tuning the model with financial-specific instructions does not impede its general generation capabilities acquired during pre-training. As a result, XuanYuan 2.0 can effectively leverage both its general-domain knowledge and domain-specific financial knowledge to provide accurate and contextually appropriate responses in the Chinese financial domain.

\begin{figure*}
\centering
\includegraphics[scale=0.6]{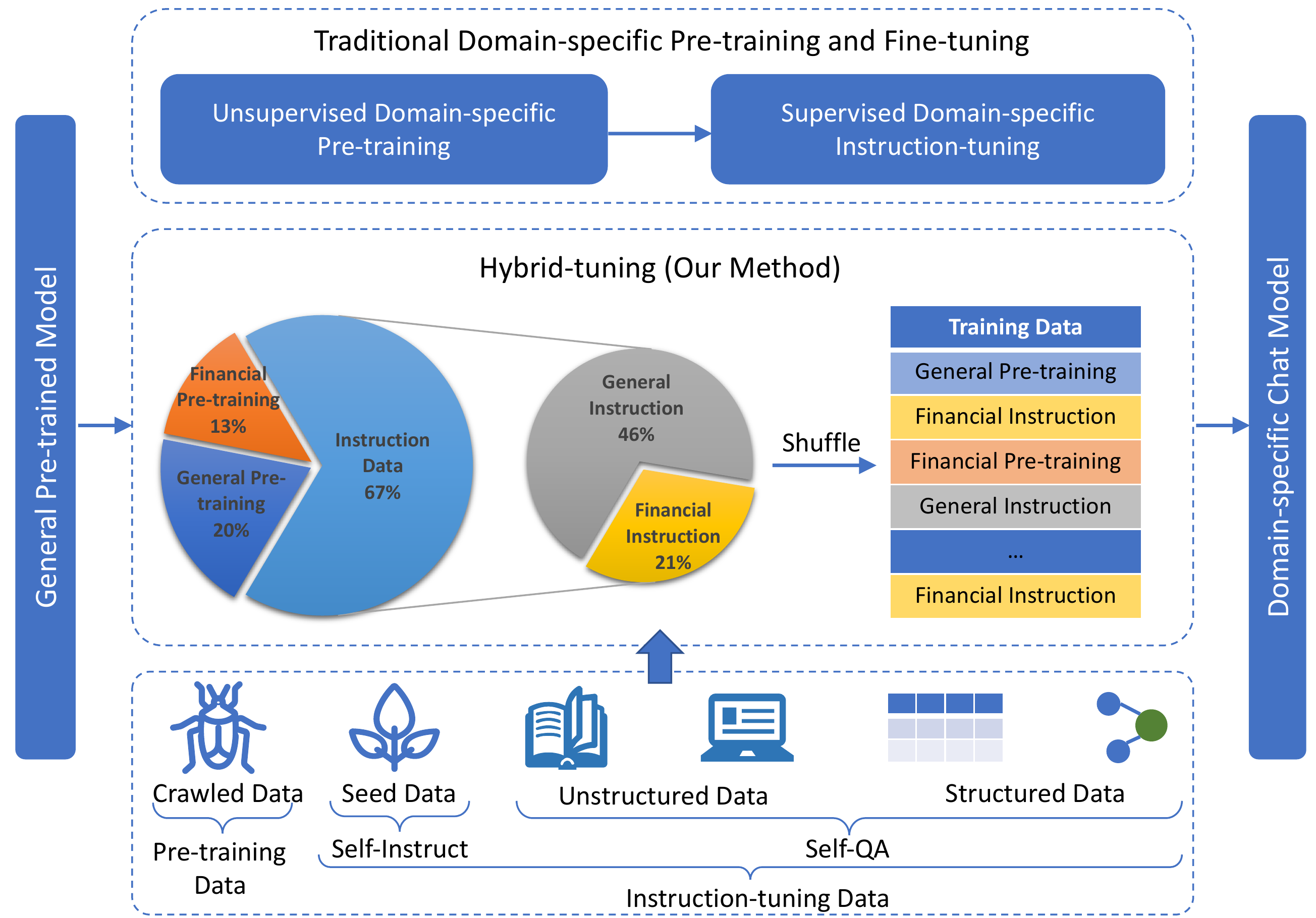}
\caption{Our proposed hybrid-tuning.}
\label{fig:method}
\end{figure*}

\section{Related Work}

The advancements in pre-trained language models have led to remarkable progress in various NLP tasks, attracting extensive research efforts. Among the notable contributions, the BERT \cite{devlin2018bert} series stands out as a groundbreaking development in the field of pre-trained models. 
Following the success of BERT, the GPT \cite{radford2018improving} series emerged as a prominent line of research, focusing on the decoding aspect of language modeling. GPT models, in contrast to BERT's bidirectional approach, leveraged autoregressive language modeling. By training on large amounts of unlabeled text data, GPT models acquired a rich understanding of language and demonstrated impressive capabilities in generating coherent and contextually relevant text. Subsequent iterations of the GPT series, such as  GPT-4 \cite{openai2023gpt4}, showcased superior performance in various language generation tasks.
And ChatGPT \cite{openai_chatgpt}, an extension of the GPT series, demonstrated the ability to engage in interactive and contextually coherent conversations. This breakthrough sparked considerable interest in developing conversational AI agents capable of simulating human-like dialogue.

In addition to the general-purpose BERT and GPT models, there has been a growing interest in domain-specific pre-training. Researchers have recognized that incorporating domain-specific knowledge during pre-training can lead to substantial performance gains in downstream tasks within those domains. Domain-specific pre-trained models aim to capture domain-specific nuances, enabling them to excel in tasks relevant to the target domain.
For instance, in the biomedical domain, BioBERT \cite{lee2020biobert} and PubMedBERT \cite{gu2021domain} are proposed to leverage large-scale biomedical corpora during pre-training. 
Similarly, in the financial domain, models such as BloombergGPT \cite{wu2023bloomberggpt} were developed to address the unique challenges and intricacies of the financial text.

Despite the advancements in domain-specific pre-training, the availability of large-scale open-source chat models specifically tailored for the Chinese language and the Chinese financial domain has remained limited. This gap motivates our work in proposing XuanYuan 2.0, a model built upon BLOOM-176B \cite{scao2022bloom} with hundreds of billions parameters, to address the unique requirements of the Chinese financial domain and facilitate the development of sophisticated conversational AI systems.

\section{XuanYuan 2.0}

\subsection{Model Architecture}

We adopted the original BLOOM \cite{scao2022bloom} architecture, which is a decoder-only architecture. The joint probability of tokens in a text can be represented as:
\begin{equation}
    p(w) = p(w_1, \ldots, w_T) = \prod_{t = 1}^T p(w_t | w_{< t})
\end{equation}
where $w$ represents a sequence of tokens, $w_t$ is the $t^{\mathrm{th}}$ token, and $w_{< t}$ is the sequence of tokens preceding $w_t$.
This method is called autoregressive language modeling, where we predict the probability of the next token in an iterative manner.
And following BLOOM, we utilize ALiBi positional embeddings \citep{press2021train} and embedding LayerNorm ~\citep{dettmers2022llm} in the traditional decoder structure of Transformer \cite{vaswani2017attention}.

\subsection{Hybrid-tuning}

To alleviate the problem of catastrophic forgetting, we propose a novel domain-specific training framework, hybrid-tuning.
In terms of the training stage, it integrates the pre-training stage and instruction fine-tuning stage that are previously split together.
In terms of the field of data, it integrates data from both general and financial domains.

As shown in Figure \ref{fig:method}, 
different from traditional two-stage domain-specific training,
our proposed hybrid-tuning randomly shuffles pre-training data (general pre-training, financial pre-training) and instruction data (general instruction, financial instruction) into one training data.
And all the training process is done in one stage.
In this way, the model can accurately handle instructions in the financial domain, while retaining general conversational capabilities.

For unsupervised pre-training data, we crawl them from the Internet and clean and filter them.
For Instruction-tuning data, we use human-written seed instructions to collect general data  by Self-Instruct \cite{wang2022self} and utilize unstructured and structured data in the financial field to gather domain-specific instruction data by Self-QA \cite{zhang2023selfqa}.
Unstructured financial data comprises a wide range of textual information, such as financial news articles, market reports, analyst commentary, and social media discussions. 
And structured financial data includes company information and so on.
These sources offer valuable insights into market trends, investment strategies, and economic situations.

\subsection{Training}
To train our complex and computationally intensive model, we employ the powerful NVIDIA A100 80GB GPU and the DeepSpeed \cite{rasley2020deepspeed} distributed training framework. For parallel processing, we primarily rely on pipeline parallelism, which involves distributing the layers of our model across several GPUs. This approach ensures that each GPU only handles a portion of the model's layers, a technique also known as vertical parallelism. Additionally, we adopt the Zero Redundancy Optimizer \cite{rajbhandari2020zero} to enable different processes to store only a portion of the data (parameters, gradients, and optimizer states). Specifically, we use ZeRO stage 1, which means that only the optimizer states are divided using this method. 
The specific hyperparameters are presented in Table \ref{tab:para}.

\section{Experiment}

We conducted a comparison between our model and other open-source Chinese conversational models. Simultaneously, we constructed evaluation datasets encompassing various dimensions in both general and financial domains, which were subsequently subject to manual assessment. The results revealed XuanYuan's robust knowledge base and conversational capabilities in the financial domain. Further insights and additional findings will be presented in the next version of the paper after the release of the evaluation rankings.

\begin{table}[htb!]
\begin{center}
\resizebox{0.5\textwidth}{!}{
    \begin{tabular}{l|c|c}
    \toprule 
    Hyperparameter  &  XuanYuan2-7B & XuanYuan2 \\ 
    \midrule
    \multicolumn{3}{c}{\emph{Architecture hyperparameters}} \\
    \midrule
    Parameters &  7,069M & 176,247M \\ 
    Layers   &  30 & 70 \\ 
    Hidden dim. &  4096 & 14336 \\ 
    Attention heads &  32 & 112 \\
    Vocab size & \multicolumn{2}{c}{250,680} \\ 
    Sequence length & \multicolumn{2}{c}{2048}\\
        Precision & \multicolumn{2}{c}{\texttt{float16}}  \\ 
    Activation & \multicolumn{2}{c}{\texttt{GELU}}  \\
    Position emb. & \multicolumn{2}{c}{\texttt{Alibi}}  \\ 
    Tied emb. & \multicolumn{2}{c}{\texttt{True}} \\ 
    \midrule
    \multicolumn{3}{c}{\emph{Pretraining hyperparameters}} \\
    \midrule
    Global Batch Size  &  512 & 2048 \\ 
    Learning rate & 1.2e-4 &  6e-5  \\ 
    Total tokens & 341B & 366B \\ 
        Min. learning rate & 1e-5 & 6e-6 \\ 
    Warmup tokens & \multicolumn{2}{c}{375M}  \\ 
    Decay tokens & \multicolumn{2}{c}{410B} \\ 
    Decay style & \multicolumn{2}{c}{\texttt{cosine}} \\ 
    Adam $(\beta_1, \beta_2)$ & \multicolumn{2}{c}{(0.9, 0.95)}  \\     
    Weight decay & \multicolumn{2}{c}{1e-1} \\ 
    Gradient clipping & \multicolumn{2}{c}{1.0}  \\
    \midrule
    \multicolumn{3}{c}{\emph{Multitask finetuning hyperparameters}} \\
    \midrule
    Global Batch Size  & 2048 & 2048 \\ 
    Learning rate  & 2.0e-5 &  2.0e-5  \\ 
    Total tokens & \multicolumn{2}{c}{13B}  \\ 
    Warmup tokens & \multicolumn{2}{c}{0}  \\ 
    Decay style & \multicolumn{2}{c}{\texttt{constant}}
    \\     
    Weight decay & \multicolumn{2}{c}{1e-4}  \\ 
    \bottomrule
    \end{tabular}
}
\caption{Training hyperparameters of XuanYuan 2.0.}
\label{tab:para}
\end{center}
\end{table}

\section{Conclusion}
In this paper, we propose the largest Chinese financial chat model, XuanYuan 2.0 (\begin{CJK}{UTF8}{gbsn}轩辕\end{CJK} 2.0), to fill the gap of open-source billion-scale chat models specifically designed for the Chinese financial domain. Besides, we propose a novel training method called hybrid-tuning to mitigate catastrophic forgetting. By combining the general domain with domain-specific knowledge and integrating the stages of pre-training and finetuning, XuanYuan 2.0 achieves the remarkable ability to deliver precise and contextually relevant responses within the Chinese financial domain.
We will continue to gather larger-scale Chinese financial domain data in order to further optimize our model.

\bibliography{anthology}
\bibliographystyle{acl_natbib}

\end{document}